# Title:

Neural Network-Based Histologic Remission Prediction In Ulcerative Colitis

## Authors' names and institutions:

**Yemin li**[a,b], **Zhongcheng Liu**[c], **Xiaoying Lou**[d], Mirigual Kurban[c], Miao Li[e,f], Jie Yang[a,b], Kaiwei Che[a,b], Jiankun Wang[a,b,g,*], Max Q.-H Meng[a,b,*], Yan Huang[d,*], Qin Guo[e,f,*], Pinjin Hu[e,f]

[a]Shenzhen Key Laboratory of Robotics Perception and Intelligence, Shenzhen, China;

[b]Department of Electronic and Electrical Engineering, Southern University of Science and Technology, Shenzhen, China;

[c]Department of Small bowel endoscopy, the Sixth Affiliated Hospital, Sun Yat-sen University, China;

[d]Department of pathology, the Sixth Affiliated Hospital, Sun Yat-sen University, China;

[e]Department of Gastroenterology, the Sixth Affiliated Hospital, Sun Yat-sen University, China;

[f]Guangdong Provincial Key Laboratory of Colorectal and Pelvic Floor Diseases, The Sixth Affiliated Hospital, Sun Yat-sen University;

[g]Jiaxing Research Institute, Southern University of Science and Technology, Jiaxing, China.

## corresponding author contact information:

∗ Corresponding author.
E-mail address: wangjk@sustech.edu.cn; max.meng@ieee.org; huangy27@mail.sysu.edu.cn; guoq83@mail.sysu.edu.cn;

## Data Availability:

This work is supported by the Shenzhen Key Laboratory of Robotics Perception and Intelligence under Grant ZDSYS20200810171800001, Shenzhen Outstanding Scientific and Technological Innovation Talents Training Project under Grant RCBS20221008093305007, and National Natural Science Foundation of China grant #62103181

# WHAT YOU NEED TO KNOW

## BACKGROUND AND CONTEXT

Histologic evaluation is considered as a new therapeutic target for the palliation of ulcerative colitis; however, it is limited by the subjective influence of pathologists and high time consumption.

## NEW FINDINGS

The proposed neural network model can rapidly complete pathology scores based on the Geboes score and agree highly with clinical results.

## LIMITATIONS

The proposed neural network model cannot reflect the severity of the Geboes score.

## CLINICAL RESEARCH RELEVANCE

The proposed neural network model can distinguish histologic remission/activity in endocytoscopy images and maintain consistency of diagnostic results, which can help accelerate clinical histological diagnosis.

## BASIC RESEARCH RELEVANCE

The proposed neural network model has the characteristics of automation, accuracy, efficiency, and scale data analysis in pathology recognition in endocytoscopy. It solves the problems of subjectivity and time efficiency existing in traditional methods and helps to improve the accuracy of pathological diagnosis, work efficiency, and patient treatment quality.

# Lay summary

We develop a specific neural network model that can distinguish histologic remission/activity in endocytoscopy images of ulcerative colitis, which helps to accelerate clinical histological diagnosis.

# Abstract

**BACKGROUND & AIMS:** Histological remission (HR) is advocated and considered as a new therapeutic target in ulcerative colitis (UC). Diagnosis of histologic remission currently relies on biopsy; during this process, patients are at risk for bleeding, infection, and post-biopsy fibrosis. In addition, histologic response scoring is complex and time-consuming, and there is heterogeneity among pathologists. Endocytoscopy (EC) is a novel ultra-high magnification endoscopic technique that can provide excellent in vivo assessment of glands. Based on the EC technique, we propose a neural network model that can assess histological disease activity in UC using EC images to address the above issues. The experiment results demonstrate that the proposed method can assist patients in precise treatment and prognostic assessment.

**METHODS:** We construct a neural network model for UC evaluation. A total of 5105 images of 154 intestinal segments from 87 patients undergoing EC treatment at a center in China between March 2022 and March 2023 are scored according to the Geboes score. Subsequently, 103 intestinal segments are used as the training set, 16 intestinal segments are used as the validation set for neural network training, and the remaining 35 intestinal segments are used as the test set to measure the model performance together with the validation set.

**RESULTS:** By treating HR as a negative category and histologic activity as a positive category, the proposed neural network model can achieve an accuracy of 0.9, a specificity of 0.95, a sensitivity of 0.75, and an area under the curve (AUC) of 0.81.

**CONCLUSION:** We develop a specific neural network model that can distinguish histologic remission/activity in EC images of UC, which helps to accelerate clinical histological diagnosis.

## Introduction

Ulcerative colitis (UC) is a chronic inflammatory bowel disease caused by genetic background, environmental, luminal factors, and mucosal immune dysregulation. UC has a high prevalence in developed countries and a substantially increased prevalence in developing countries, thus it has become a global burden. UC is now considered as a progressive disease due to the risks of proximal extension, strictures, gut dysmotility, anorectal dysfunction, need for colectomy, hospitalization, colorectal cancer, disability, and reduced quality of life. Due to its potentially progressive and debilitating course, the treatment goals for UC have changed from treating symptoms to histologic remission (HR) over the past decade, aiming to modify the natural history of the disease and improve the long-term prognosis.[1] HR is critical for determining UC inflammatory activity and monitoring treatment response. It is an emerging therapeutic target and an important outcome in UC clinical trials.

The importance of HR has been known since 1966 when Wright and Truelove performed serial biopsies on UC patients and concluded that patients in HR are more frequently in clinical remission after 1 year (40% vs. 18%).[2] Since then, several clinical trials have reported various outcomes of UC and the influence of some intervening factors (particularly HR). A common concern in longstanding patients with UC is the increased risk of dysplasia and colorectal cancer, which is thought as a result of persistent colonic inflammation. Researchers have reported that histologic activity patients have a higher risk of developing dysplasia and progressing to colorectal cancer than HR patients.[3] More recently, it has been found that HR also associates with a reduced risk of hospitalization and colectomy at up to 6 years of follow-up.[4-6] However, challenges remain in integrating histology into clinical practice. Over the past few decades, more than 30 histologic scoring methods have been developed, of which the Geboes score and the Riley score have made it possible to measure histology objectively because of good inter-observer agreement.[7] However, their use in clinical practice still needs to be improved.

For histologic evaluation, biopsy is essential.[8] However, this method has obvious limitations, such as being constrained by the time required for final diagnosis, additional costs, limited sample size, risk of bleeding, sampling errors, and the possibility of inducing congenital mucosal breaks and submucosal fibrosis, which may hamper subsequent endoscopic therapies.[9] In addition, histological assessments require training, and all scoring systems are operator-dependent, with varying interobserver agreement. Even experienced specialists have varying interobserver agreement. Although training programs can improve consistency, subjectivity hinders its use in clinical practice.

In recent years, several "optical biopsy" techniques have been introduced, attracting widespread attention and scientific advances.[9] As one of them, EC is a novel ultra-high magnification endoscopic technique that allows high-quality in vivo evaluation of lesions found in the gastrointestinal tract by intraoperative staining.[10] It allows non-invasive visualization of mucosal cells and glandular structures and real-time in vivo visualization of nuclei and microvasculature, facilitating "optical biopsy" or "virtual histology" of colorectal polyps or tumors.[11] It has significantly advanced diagnosing ulcerative colitis by evaluating inflammatory activities such as crypt, cellular infiltration, and microvascular structures.

Advances in Artificial intelligence (AI) technologies have yielded exciting results in replicating expert judgment and predicting clinical outcomes, particularly in the analysis of imaging.[12-15] For example, validation studies in multiple sclerosis have demonstrated the ability of AI to replicate expert assessment of static cellular images; Ozawa et al. develop a computer-aided diagnostic system using colonoscopic static images that excels in differentiating between Mayo Endoscopy Score (MES) remission (MES 0,1) and active disease (MES 2,3, ROC 0.98).[16] AI directly affects how we assess, monitor and manage inflammatory bowel disease, and here for the first time, we use endoscopic images to evaluate HR. We build a neural network that takes EC images as inputs and clinical diagnostic results as labels, and by using supervised learning, it can utilize the full information of the intestinal segments to complete the pathology diagnosis rapidly. The method optimizes the influence of local information on the results and improves the accuracy of diagnosis; it can free doctors from tedious work and help balance medical resources.

## Methods
### 1. Patients
We screen patients with long-standing UC undergoing endoscopic examination at a hospital between March 2022 and March 2023. The exclusion criteria are as follows (1) patients with previous colon surgery, unclassified inflammatory bowel disease, Crohn's disease, colorectal neoplasia, or co-infectious colitis (examined at the time of blood and stool testing); (2) patients with contraindications to colonoscopy; and (3) patients with contraindications to biopsy due to hematopoietic disorders or antithrombotic, anticoagulant therapy. All patients are given a standard bowel preparation. We use the Montreal criteria to determine the degree of disease.[17] And the quality of colon cleansing is recorded.

## 2. Endoscopic evaluation

We perform standard colonoscopies (Olympus EC 520-fold), capture endoscopic images of each colonic segment (ascending, transverse, descending, sigmoid, and rectum), and subsequently stain the most severe area with crystal violet and methylene blue: 0.05% crystal violet (CV) +1% methylene blue (MB) through the standard spray tube (Olympus), and the staining time is 90 seconds to 180 seconds. Then we observe the mucous with the magnification of EC up to 520 times (Olympus EC 520-fold). At least five clear endoscopic images should be taken of each bowel segment, and then we obtain two mucosal biopsy samples from the center of the corresponding region.

## 3. Histological evaluation

Formalin-fixed paraffin-embedded biopsy tissue blocks are cut into 4-μm sections for hematoxylin and eosin staining. Histological inflammatory activity is scored using the Geboes score by experienced gastrointestinal pathologists (Dr. YH and Dr. XYL). It is divided into 6 grades: architectural changes [Grade 0], chronic inflammatory infiltrate [Grade 1], lamina propria neutrophils and eosinophils [Grade 2], neutrophils in the epithelium [Grade 3], crypt destruction [Grade 4], and erosions or ulcerations [Grade 5], and each grade of the score is divided into four subcategories. Geboes grade less than 3.2 is defined as histological remission. The specific manifestations are shown in [Figure 1].

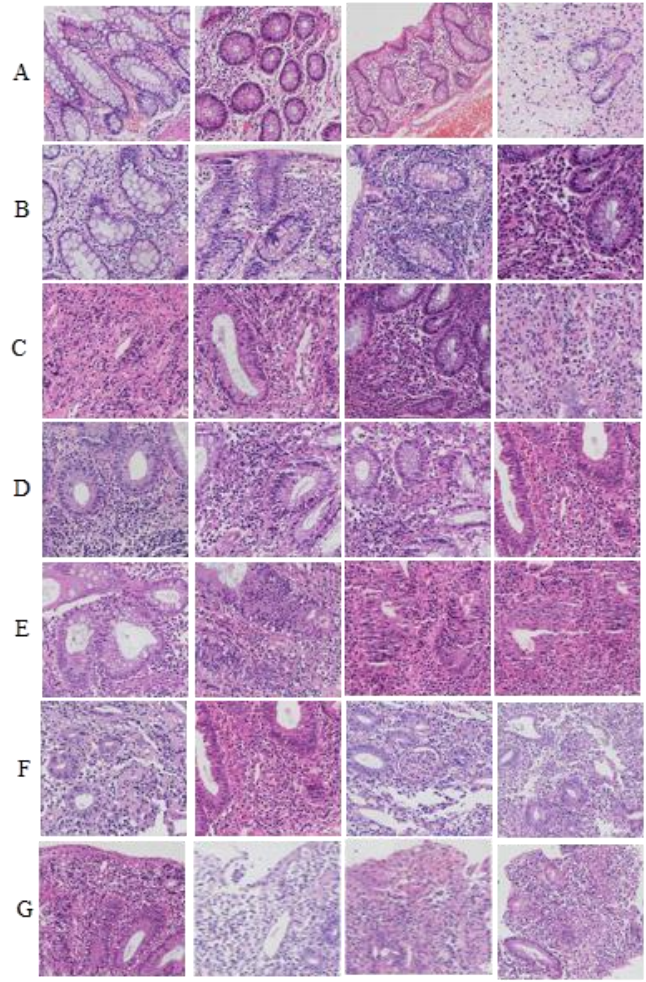

**Figure 1.** Histological inflammatory activity is scored using the Geboes score: (A) Grade 0; (B) Grade 1; (C) Grade 2A; (D) Grade 2B; (E) Grade 3; (F) Grade 4; (G) Grade 5.

## 4. The proposed neural network model

To accomplish the task of rapid determination of whether a patient's UC is HR, we design different networks for testing. The used datasets are obtained from 154 bowel segment images and clinical diagnostic results. For dataset preparation, we clean, enhance, and resample the existing dataset and then feed it into the network for training and testing. For network design, we first use Residual Network (ResNet) and Vision Transformer (ViT) as the backbone for feature extraction and then use multilayer perceptron for classification. Based on the experimental results of this part, we modify the existing network by implementing the attention mechanism in ResNet (ResNet-A) [Figure 2], which can both strengthen the feature representation and reduce the model parameters and computational complexity. The results indicate an improvement in model performance after adding the attention mechanism. Ultimately, we design our network, which uses the attention mechanism for feature extraction, uses the residual connection to optimize the model convergence, and finally completes the classification and result output [Figure 3]. For the training policies, we compare the effects of different input image sizes ([224, 224], [299, 299], [512, 512]) and resampling strategies (Synthetic Minority Over-Sampling Technique (SMOTE), Random Under-sample and Oversample (RUAO)) on the results and choose the optimal strategy for subsequent training. At last, we evaluate the model's performance by comparing the model's prediction results in the test set with the clinical results. For more information about the model, please refer to the Supplementary Material.

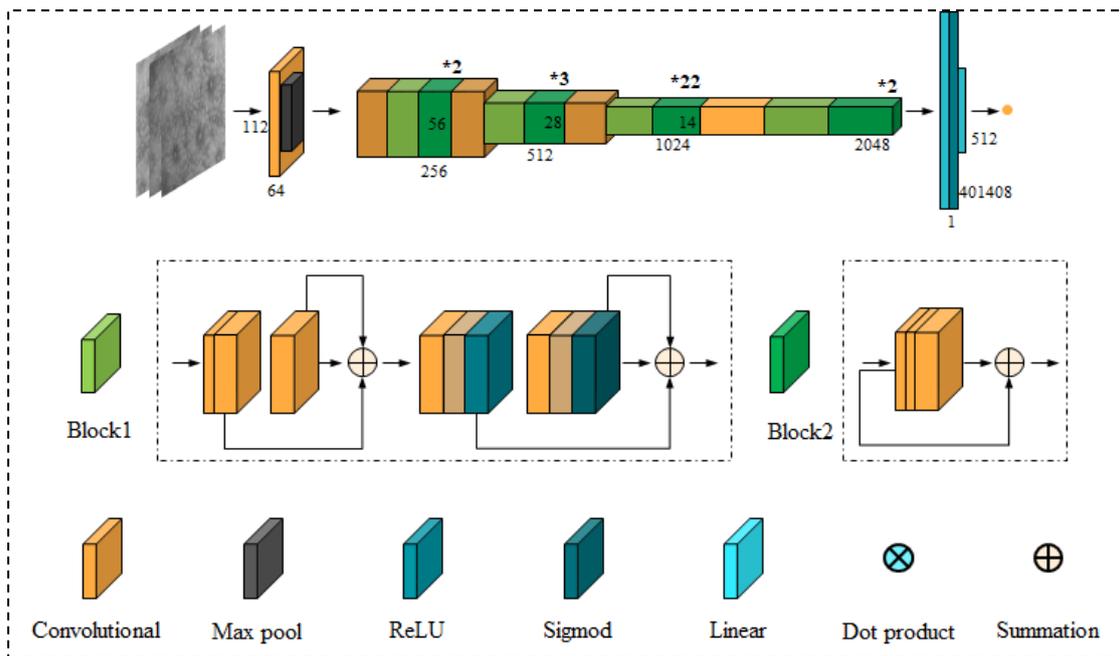

**Figure 2.** ResNet-A model.

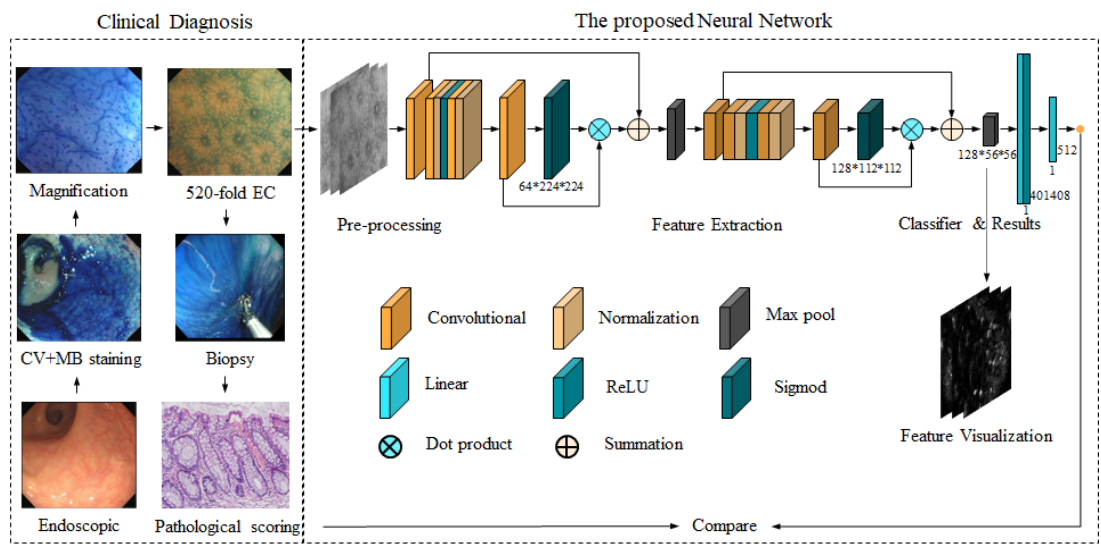

**Figure 3.** The clinical diagnostic process and our proposed model.

## 5. Evaluation metrics

By treating histologic remission as a negative category and histologic activity as a positive category, the diagnostic metrics used by the proposed neural network to classify UC remission/activity include accuracy, sensitivity, specificity, and area under the receiver operating characteristic curve (AUC). Among them, accuracy indicates the percentage of correctly predicted samples to the whole test set; sensitivity indicates the percentage of correctly predicted positive samples to the overall positive samples; specificity indicates the percentage of correctly predicted negative samples to the overall negative samples; and AUC is used to measure the performance of the model intuitively.

## Results

### 1. Patients

Excluding five patients with unclear images and six patients with severe Ulcerative colitis accompanied by gastrointestinal bleeding, 87 patients with Ulcerative colitis are included in the study. 326 mucosal tissue pathologies are included, including 5105 EC images. 50.6% of the patients are male, 49.4% are female, the mean age is 40±15.07, and the average course of the disease is 1.7 ± 4.5. The monthly criteria are E1 32.2%, E2 5.7%, E3 62.1%, MES are Mayo 0 76.9%, Mayo 1 11.5%, Mayo 2 6.9%, Mayo 3 5.5%. Patients receive treatments including Mesalazine 62.1%, Steroids 5.75%, Vedolizumab 22.9%, Ustekinumab 1.1%, and Infliximab 8.0%.

**Table 1.** Patient situation

|  |  |
|---|---|
| Sex | F:44, M:43 (F:50.6%, M:49.4%) |
| Mean age ± SD (range) | 40±15.07(17-78) |
| Disease duration median years (range) | 1.7±4.5(0-30) |
| Localization |  |
| E1 | 28(32.2%) |
| E2 | 5(5.7%) |
| E3 | 54(62.1%) |
| Mayo endoscopy score |  |
| Mayo 0 | 66(76.9%) |
| Mayo 1 | 10 (11.5%) |
| Mayo 2 | 6 (6.9%) |
| Mayo 3 | 5 (5.5%) |
| Medication | |
| Mesalazine | 54(62.1%) |
| Steroids | 5 (5.75%) |
| Vedolizumab | 20 (22.9%) |
| Ustekinumab | 1(1.1%) |
| Infliximab | 7(8.0%) |

## 2. Mucosal healing and histological remission

Colonoscopy can visualize the mucosal surface of the bowel and is used to assess significant mucosal inflammation. All patients need to undergo a complete colonoscopy and will be scored according to the Mayo score. However, mucosal healing observed by colonoscopy is not representative of histologic remission [Figure 4]. Double staining with CV and MB makes the nuclei and glands on the surface of the intestinal mucosa recognizable by endoscopy, making endoscopic determination of HR possible [Figure 5].

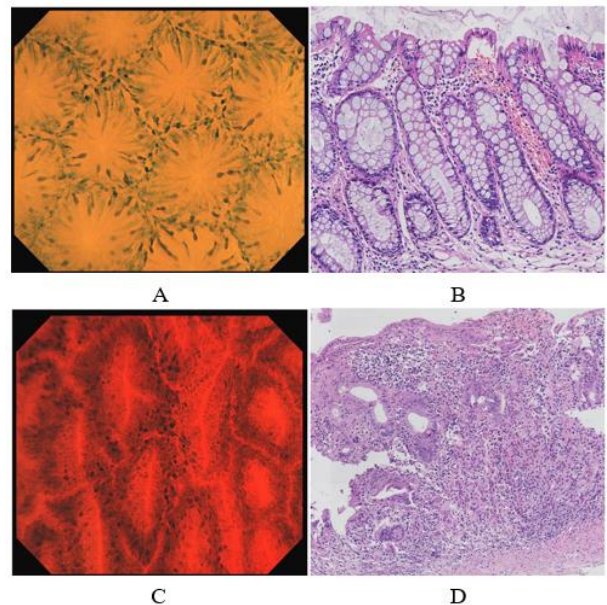

**Figure 4.** Examples of the mucosal healing and non-mucosal healing. (A) EC images of mucosal healing; (B) Pathological section of corresponding area; (C) EC images of non-mucosal healing; (D) Pathological section of corresponding area.

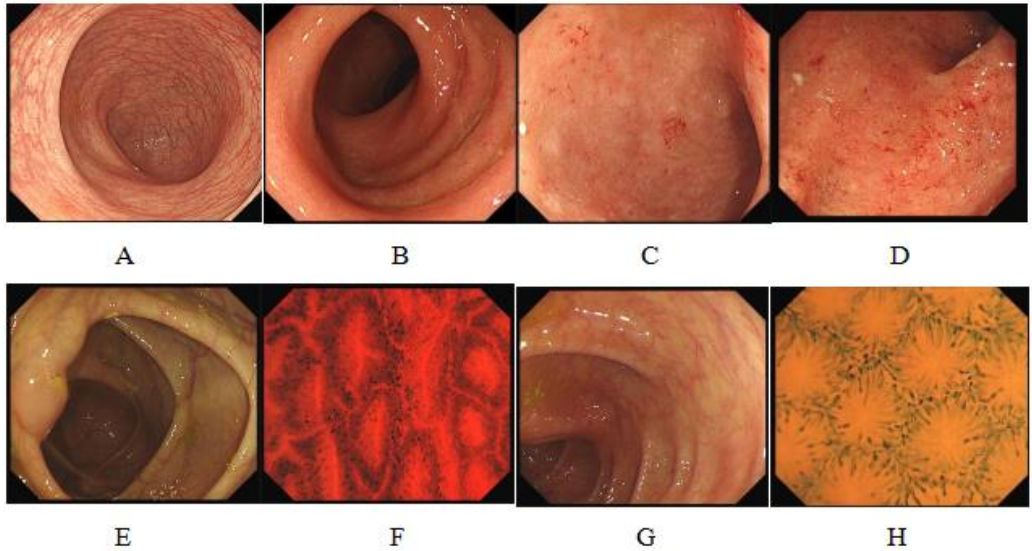

**Figure 5.** EC images from Mayo score 0 to 3: (A) Mayo 0, (B) Mayo 1, (C) Mayo 2, (D) Mayo 3; (E) Colonoscopic images of non-mucosal healing of ascending colon; (F) EC images of non-mucosal healing; (G) Colonoscopic images of mucosal healing of ascending colon; (H) EC images of mucosal healing.

## 3. Results of the neural network model

When analyzing EC images clinically, we will focus on the following image features: (1) whether the opening of the glandular duct is standard, (2) whether the glandular interspace is normal, (3) whether there is the phenomenon of inflammatory cell infiltration. In contrast to manual analysis by pathologists, Convolutional Neural Networks (CNNs) play an essential role in medical image analysis due to their fast, multidimensional learning and interpretation of image information.

To solve the task, we design and validate several networks. In the first part of the experiments, we use ResNet-101 and EfficientNet-B0 as the backbone network to extract image features, respectively. Afterward, we perform the downscaling and result output through the fully connected layer. Meanwhile, we compare the effects of different input image sizes and resampling strategies.

In EC images, pathological manifestations such as inflammatory cell infiltration are often discontinuous, but the convolutional layer in traditional CNNs usually can only capture local information. To make our network able to capture global pathological features, we test the Transformer network in the hope of improving the network performance by introducing the attention mechanism. In this part of the experiment, we use ViT-Base and ViT-Large as the backbone network, respectively, and select the better strategy according to the first part of the experiment. Considering that the attention mechanism in ViT networks is designed for textual content, the performance in image tasks still needs to be improved. To verify the role of this mechanism in the network, we add the image-based attention mechanism to the hierarchical design of the ResNet and select the better performing strategies for training. The results of this part of the experiment are shown in [Table 2].

**Table 2.** Results of the neural network model

| ID  | Backbone        | Image size  | Resampling | Accuracy | Sensitivity | Specificity | AUC  |
| --- | --------------- | ----------- | ---------- | -------- | ----------- | ----------- | ---- |
| 1.1 | ResNet-101      | [224, 224]  | RUAO       | 0.78     | 0.5         | 0.87        | 0.7  |
| 1.2 | ResNet-101      | [299, 299]  | RUAO       | 0.75     | 0.5         | 0.82        | 0.65 |
| 1.3 | ResNet-101      | [512, 512]  | RUAO       | 0.75     | 0.42        | 0.85        | 0.67 |
| 1.4 | ResNet-101      | [224, 224]  | NO         | 0.75     | 0.58        | 0.8         | 0.66 |
| 1.5 | EfficientNet-B0 | [224, 224]  | NO         | 0.75     | 0.5         | 0.82        | 0.65 |
| 1.6 | ResNet-101      | [224, 224]  | RUAO       | 0.82     | 0.58        | 0.9         | 0.76 |
| 1.7 | EfficientNet-B0 | [224, 224]  | RUAO       | 0.86     | 0.67        | 0.92        | 0.77 |
| 1.8 | ResNet-101      | [224, 224]  | SMOTE      | 0.75     | 0.42        | 0.85        | 0.68 |
| 1.9 | EfficientNet-B0 | [224, 224]  | SMOTE      | 0.82     | 0.58        | 0.9         | 0.73 |
| 2.1 | ViT-Base        | [224, 224]  | RUAO       | 0.59     | 0.33        | 0.67        | 0.55 |
| 2.2 | ViT-Large       | [224, 224]  | RUAO       | 0.61     | 0.33        | 0.69        | 0.54 |
| 3.1 | **ResNet-A**    | [224, 224]  | RUAO       | 0.84     | 0.41        | 0.91        | 0.76 |
| 3.2 | **Our**         | [224, 224]  | RUAO       | 0.9      | 0.75        | 0.95        | 0.81 |

As can be seen from the table, based on the fixed backbone network and resampling method, using three common image sizes as input, the conclusion is that this variable does not contribute significantly to the performance improvement of the model. The possible reason is that the EC images have highly similar content and structure; thus, the different sizes of input images will not affect the results effectively.

Then with fixed input image size and backbone network, the results of no resampling, using SMOTE and resampling using RUAO, are separately evaluated. The results show that the performance of the model with resampling is improved compared to no resampling, which means that when the sample data are unbalanced, the appropriate resampling method can help us optimize the model performance. After using the resampling method, the RUAO method performs better than the SMOTE method. The analytical reason is that SMOTE uses the sample data for feature extraction and synthesis. However, for EC images, the detail part of the image is the key to determining the output. When performing feature synthesis, it is impossible to assess whether the synthesized features satisfy the pathological conditions precisely; hence its result is less efficient than directly oversampling the pathological images.

Based on the former experiments, we fix the input image size and use the more effective resampling method to verify the model performance with different transformer structures as the backbone. We can observe that the system's overall performance after using the transformer structure is insufficient for the model trained with the CNN. The reason is that the Transformer network usually needs more datasets for training to achieve a better result. However, in EC dataset, the sample amount is small. Even after dataset resampling and data augmentation, the model cannot converge. Meanwhile, it can be concluded from the results that the ResNet-A has some improvement compared to the original transformer network, so it is feasible to use the improved method of using an image-based attention mechanism and reducing the number of network parameters.

To better utilize the attention mechanism to help us capture the feature information in the image, we designed a new network to test. The new network uses the attention mechanism to extract feature information from the image, uses residual connectivity to optimize convergence, and then passes the extracted features through a multilayer perceptron for classification. The final experimental results are also shown in [Table 2]. Compared with the former results, the model has better performance. Some test set images and their corresponding predicted and accurate results are shown in [Table 3].

**Table 3.** Pathologic images with their corresponding predicted and clinical results

| ID | EC images | Predicted results | Clinical results |
|----|-----------|-------------------|------------------|
| 1  | 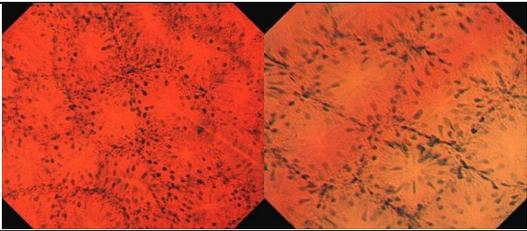 | 0 | 0 |
| 2  | 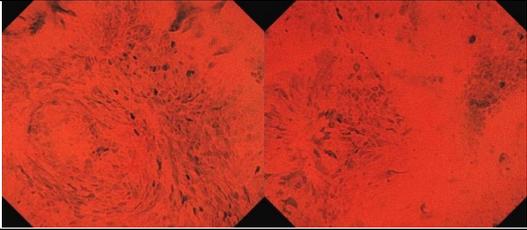 | 1 | 1 |

## Discussion

In this work, our contributions can be summarized as follows: (1) For the first time, we collect and organize datasets from different bowel segments of UC patients, including white light images, EC images, and clinical diagnostic reports, which can be used for subsequent model development and task implementation; (2) For the first time, we use neural networks to process EC images, and the model predictions are in high agreement with clinical results and (3) This work provides a reference for intelligent diagnosis of other conditions and further demonstrates the great potential of neural networks in clinical diagnosis. The use of the neural network method to assist UC diagnosis can exclude the influence of biopsy local information on the results and quickly complete the resulting output, which not only improves the consistency of clinical diagnostic results but also liberates doctors from tedious work, and helps to optimize the medical environment and balance medical resources.

The experimental results show that the model has an accuracy of 0.9, a sensitivity of 0.75, a specificity of 0.95, and an AUC of 0.81. i.e., the model can predict the majority of normal samples and more accurately predict abnormal samples, which can help doctors in the clinic to formulate a strategy for the patient's medical treatment quickly. The future direction is to refine the output of the network so that it can output specific scores for each item according to clinical criteria, i.e., it can not only determine whether the patient has histologically healed but also determine the current condition of the non-healing patient, in order to improve the diagnostic process further.

[17]Satsangi J, Silverberg MS, Vermeire S, et al. The Montreal classification of inflammatory bowel disease: controversies, consensus, and implications. Gut. 2006 Jun;55(6):749-53.

Author names in bold designate shared co-first authorship.

## Appendix

In clinical practice, it takes much time to determine whether a patient's ulcerative colitis is histologic remission; to solve this problem, we train neural networks for assisted diagnosis. On the dataset, we randomly divided 154 patient samples into 103 cases as the training set, 16 as the validation set, and 35 as the test set al. In order to better utilize the data to complete the model training, we need to perform appropriate pre-processing of the image so that the processed dataset can fulfill the following functions: (1) Excluding the effects of inconsistencies in color due to manual manipulations such as staining; (2) Excluding the effects of angular differences caused by device imaging, manual manipulation, etc.; (3) Excluding the effect of noise information.

The number of EC images collected in the clinic is small and unevenly distributed. To solve this problem, we compare two resampling methods during the dataset loading process. SMOTE can balance the number of samples in each category by synthesizing the few category samples. The specific idea is that we will count the number of each kind of sample in the preparation of the dataset and use a convolutional neural network to extract features for the image samples with fewer data, then use the extracted features to synthesize the new features, at last, the new features will be restored to the image as input for training the model; RUAO is a combination of random under-sampling and random oversampling strategy. The specific idea is that we will count the number of each sample in the preparation of the dataset, and we will randomly under-sampling for the larger samples and use random oversampling for the smaller samples.

Before loading the dataset for model training, we must match the images with the labels. Since we get patient-level labels when processing the data, we need to make all images of the patient samples share the patient labels and train the model at the image level. However, even for patient samples, the images acquired may be standard; therefore, after image-level classification, we must cluster the output results and select the highest frequency results for the final patient-level output.

After preprocessing, we can deliver the data to the neural network for training. The neural network is implemented based on torch 1.13.1 and python 3.8, and trained on NVIDIA A100 with a batch size of 128, a resampling strategy of RUAO, a learning rate of 0.001, and a loss function of cross-entropy loss. The network is trained for 400 rounds and reaches the optimization at round 245.